\documentclass[journal]{IEEEtai}

\usepackage[colorlinks,urlcolor=blue,linkcolor=blue,citecolor=blue]{hyperref}

\usepackage{color,array}
\usepackage{mathtools}

\usepackage{amsmath,amssymb,amsfonts}
\usepackage{algorithmic}
\usepackage{graphicx}
\usepackage{textcomp}
\usepackage{xcolor}
\usepackage{tikz}
\usepackage{tipa}
\usepackage{here}
\usepackage{adjustbox}
\usepackage{booktabs}
\usepackage{multirow}

\begin{document}

\title{GAN-SLAM: Real-Time GAN Aided Floor Plan Creation Through SLAM}

\author{Leon Davies, Baihua Li, Simon Sølvsten, Qinggang Meng}


\maketitle

\begin{abstract}
SLAM is a fundamental component of modern autonomous systems, providing robots and their operators with a deeper understanding of their environment. SLAM systems often encounter challenges due to the dynamic nature of robotic motion, leading to inaccuracies in mapping quality, particularly in 2D representations such as Occupancy Grid Maps. These errors can significantly degrade map quality, hindering the effectiveness of specific downstream tasks such as floor plan creation.
To address this challenge, we introduce our novel `GAN-SLAM` a new SLAM approach that leverages Generative Adversarial Networks to clean and complete occupancy grids during the SLAM process, reducing the impact of noise and inaccuracies introduced on the output map.
We adapt and integrate accurate pose estimation techniques typically used for 3D SLAM into a 2D form. This enables the quality improvement 3D LiDAR-odometry has seen in recent years to be effective for 2D representations.
Our results demonstrate substantial improvements in map fidelity and quality, with minimal noise and errors, affirming the effectiveness of GAN-SLAM in for real-world mapping applications within large scale complex environments.
We validate our approach on real-world data operating in real-time, and on famous examples of 2D maps. The improved quality of the output map enables new downstream tasks, such as floor plan drafting, further enhancing the capabilities of autonomous systems. 
Our novel approach to SLAM offers a significant step forward in the field, improving the usability for SLAM in mapping based tasks, and offers insight into the usage of GANs for OGM error correction.
\end{abstract}

\begin{IEEEImpStatement}
Occcupancy grid maps in their current form can be interpreted as unappealing to human users as their intended use is primarily  for navigation tasks by robots. 

We propose a GAN model embedded into the SLAM process trained to convert the noisy maps produced by SLAM into clean variants by identifying errors and noise in the SLAM process. 

Reliable reduction of errors and artefacts in maps produced by SLAM has the potential to improve the usability of occupancy grid mapping for new domains such as facilities management, or the insurance industry that could benefit from quickly produced accurate maps. 

Our GAN-SLAM also offers insight into reliable error reduction to common SLAM errors such as sensor noise, linear and angular offsets through GANs.

\end{IEEEImpStatement}


\section{Introduction}

\IEEEPARstart{O}{ccupancy} grid mapping (OGM) is a 2D representation of SLAM in which the world is represented as a grid map. Each grid cell represents a state of occupation for its position in the map as occupied, free or unexplored. By combining data collected from various sensors such as LiDARs or Inertial Measurement Unit (IMU) sensors, OGM enables robots to navigate and interact autonomously and dynamically with their surroundings. This enables OGMs to be a reliable format for further tasks such as exploration, obstacle avoidance and path planning. 

OGMs built through SLAM are prone to noise and artefacts due to the fundamental underlying processes of estimation which enable SLAM. 
This can be attributed to the sensors employed, such as high dimensional LiDAR scanners, environmental factors, such as the occurrence of transparent or reflective materials, and the dynamic motion of the robot itself. Additionally, errors in pose estimation and drifts in odometry contribute to inaccuracies in the generated maps such as linear or angular offsets. 
These errors may be more pronounced when utilising lower cost sensors \cite{b2}. However, higher power sensors have the potential to introduce different types of erroneous data such as LiDAR capturing data through open doorways or glass. Furthermore, Partially made observations or blocked segments of rooms due to occlusions leave unfinished sections of maps. In the past these factors have made OGM not as reliable for tasks such as automated floor plan creation.


\begin{figure}[t]
    \centering
    \includegraphics[width=1\linewidth]{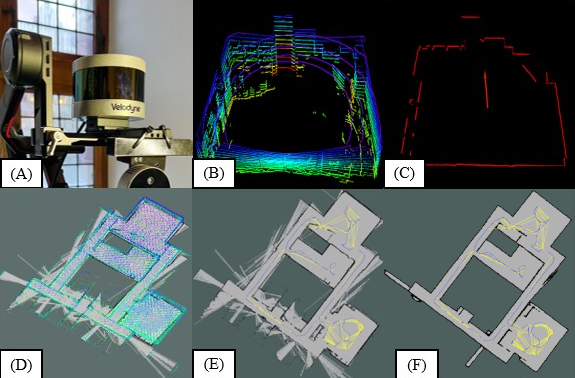}
    \caption{GAN-SLAM: Data pipeline, producing a high quality defect free OGM from the singular input of 3D LiDAR. (A) 360-degree LiDAR scanner. (B) 3D PointCloud captured by LiDAR scanner. (C) 2D LiDAR data + Odometry calculated from LiDAR data. (D) 2D/3D SLAM produced by LiDAR \& pose estimate. (E) Estimated 2D OGM. (F) GAN-Cleaned OGM.}
    \label{fig:slam_diagram}
\end{figure}

Traditional methods of SLAM optimisation and error correction include the usage of bundle adjustment \cite{b3}, loop closure \cite{b4} and probabilistic methods \cite{b6,b7}. These are fundamental processes in modern day SLAM systems. These methods are not only susceptible to irregularities and artefacts but in many cases are the root cause for these errors. Many irregularities have become accepted as normal within modern day 2D OGM. This degradation in map quality can make OGM unappealing for usage as floor plan diagrams.

Errors such as linear/angular offsets, accidentally captured observations, and partial observation completion can be difficult to resolve with traditional programming practices but are easy for humans to identify.

This inspires our proposed GAN-SLAM method which uses a deep learning approach to resolve these issues. We present a novel concept within SLAM which utilises deep neural networks trained to identify various SLAM errors, predict a completed room from a partial observation, and apply these changes and resolve them in real-time.

We implement state of the art LiDAR-odometry traditionally used in 3D SLAM to 2D OGM. This massively improves localisation accuracy within large scale complex environments, with no further sensors required. This also allows us to benefit from the increase in range that more powerful 3D scanners offer.   

Our results demonstrate robust and reliable mapping quality and offer a novel concept of floor plan creation through error correction and observation completion in OGM through GANs. Through our SLAM, complex environments are able to be successfully mapped, our method out performs all other pure LiDAR OGM SLAM tested in terms of mapping quality and accuracy.  

Mature 2D laser scan matching methods with 2D LiDARs are commonly used for OGM tasks. Our specific context involves mapping several thousand large factories across the UK with autonomous systems. In this scenario the increased range 3D LiDARs offer and more precise pose estimate is necessary. With our method we are able to produce reliable results from partially complete scans that are completed by the GAN. This is a never before seen feature within a SLAM and can amount to much better efficiency of mapping. SLAM algorithms often show success in small scale controlled environments but often fail to produce reasonable results in larger more unpredictable environments \cite{slam-errors}. With our method we prove and contribute a new SLAM that can be relied upon for large scale mapping tasks.

\section{Related Work}

\subsection{SLAM Optimisation}
Traditional methods of particle filter based SLAM enable real-time simultaneous pose estimation and map construction. Methods such as Monte Carlo Localisation \cite{b6} and Rao-Blackwellized filtering \cite{b7} utilise a probabilistic interpretation where pose is represented by a set of weighted particles, enabling robust SLAM in small-scale dynamic and unknown environments. Extended Karlman Filtering \cite{karlman} incorporates elements of Karlman filters for state estimation typically in conjunction with feature-based SLAM. These SLAM optimisation techniques may introduce errors in complex or uncertain environments and cannot discern between intended mapping and accidental mapping such as observations captured through open doorways or transparent materials.

Feature matching techniques such as the iterative closest point (ICP) algorithm \cite{b10} and newer variants \cite{b11} are standardised techniques to align pointclouds and estimate pose through optimising distances between iterative scans. 

\subsection{SLAM Algorithms}

Traditional 2D LiDAR-odometry SLAM algorithms such as Gmapping \cite{b7}, Hector-SLAM \cite{b28}, and SLAM-toolbox \cite{b29}  are widely used due to the computational efficiency and effectiveness in structured environments. However, these methods struggle with cumulative drift of odometry, inconsistent loop closures and inability to differ between correct and erroneous mapping in complex dynamic scenes.

The most influential 2D LiDAR-odometry SLAM algorithms include Gmapping \cite{b7}, Hector-SLAM \cite{b28}, and SLAM-toolbox \cite{b29}. These offer reliable results in small scale scenes on steady robots but struggle to maintain accurate pose in larger more complex environments. Making errors such as drifted odometry, inconsistency in loop closures and general failure in feature-sparse areas common \cite{slam-errors}. Evidence based occupancy mapping techniques \cite{32,33,34} are the classical and most popular method of occupancy grid mapping. 3D LiDAR-odometry algorithms such as LOAM \cite{b12} are far better suited for SLAM in complex situations, these however are not optimised to produce OGMs.

The LOAM algorithm and its various adaptations are widely used techniques for SLAM within multi-sensor loosely coupled robotic systems. Closely coupled multi-sensor systems including Direct Lidar Odometry (DLO) \cite{b17} fuse IMU with LiDAR to produce 3D maps and pose estimation. These methods require the usage of multiple sensors and are applied to 3D SLAM. We demonstrate high quality 2D mapping with the usage of a singular LiDAR scanner. None of the aforementioned methods have the ability to assume a completed observation from a partially mapped location, or removing accidentally mapped regions, which GAN-SLAM is able to do. 

\subsection{SLAM Error Correction}
Loop Closure \cite{b4} and Bundle Adjustment \cite{b3} are time tested methods which can improve the quality of a SLAM through correcting accumulated localisation errors and refining trajectory based on historical observations. Our GAN-SLAM does not attempt to replace these but can be used in conjunction with.

\subsection{Deep Learning and SLAM}
Deep learning has been applied within SLAM and has shown success in many situations. Superglue \cite{b18} is perhaps the most well known example and is applied for pose estimation. The method demonstrated in \cite{b19} proposes the usage of learning based models for loop closure and noise reduction in 3D SLAM. These methods have not been attempted for 2D OGM. Our research is the first work that combines GANs with OGM based SLAM for improved map quality and completeness through error removal and observation completion. This enables the novel application of large scale floor plan creation through OGM.


\section{Method}

We represent OGM as a matrix and denote it as $M$, where each element $M_{ij}$ of the matrix represents the occupancy status or likelihood of occupancy at grid cell $(i, j)$.

$M = \{ M_{ij} \mid M_{ij} \in [0,1] \text{ for } 1 \leq i \leq H, 1 \leq j \leq W \}$
In which
 $M_{ij} \approx 0$ indicates high confidence that cell $(i, j)$ is unoccupied and,
 $M_{ij} \approx 1$ indicates high confidence that cell $(i, j)$ is occupied. $H$ and $W$ represent dimensions Height, and Width.

The data format of an OGM can be represented as an OGM[$H \times W$] akin to a matrix [$H \times W$] in structure. We use this to justify the adaptation of deep learning methodologies, traditionally applied to image data, for occupancy grid maps. 
More specifically we treat the task of occupancy grid cleaning and artefact removal as an image-to-image (I2I), or OGM-to-OGM translation task, and train a GAN to learn this mapping. 
The objective of an I2I GAN is to translate images  from input domain $X \subset \mathbb{R}^{H \times W \times 3}$
to appear like images in output domain $Y \subset \mathbb{R}^{H \times W \times 3}$. 
In our case domain $X$ refers to erroneous  OGMs, and domain $Y$ refers to pixel perfect OGMs.

\subsection{GAN-SLAM System Architecture}

\begin{figure*}[htbp]
    \centering
    \includegraphics[width=\textwidth]{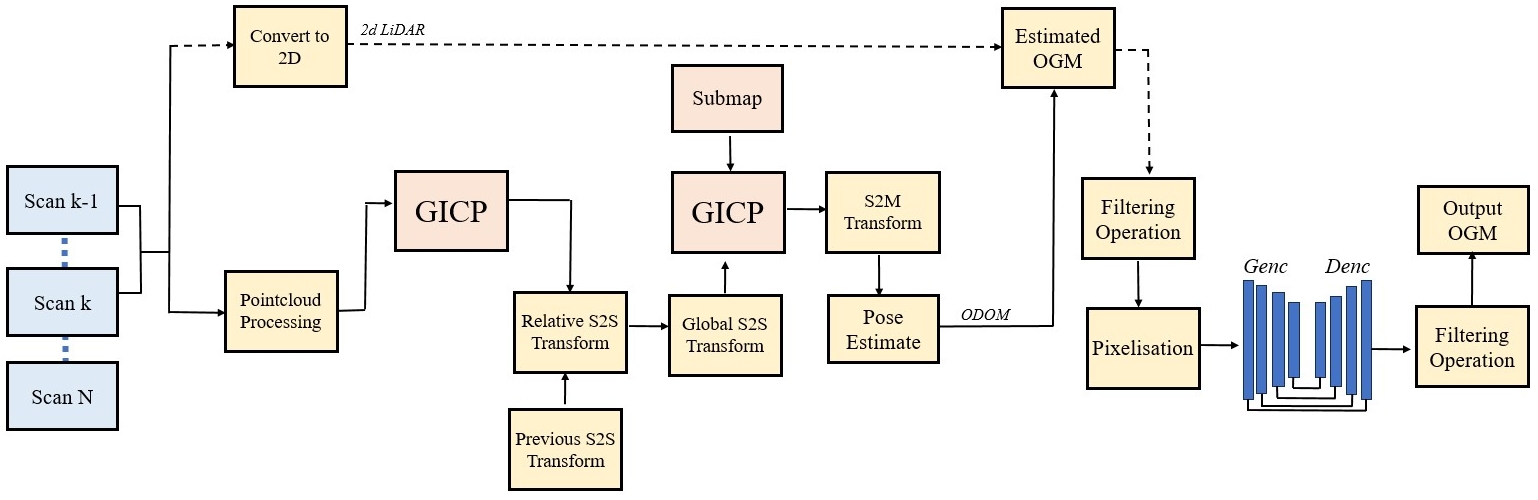}
    \caption{GAN-SLAM System Architecture Diagram for real-time occupancy grid mapping. 
    Main stages: 1) Take the input of consecutive LiDAR pointclouds to calculate pose through generalised-ICP and \cite{b17}. 
    2) Combine calculated pose with 2D conversion of incoming scan to build an estimated $OGM_e$. 3) Preprocess $OGM_e$ through filtering operation and pixelisation. 4) Parse preprocessed $OGM_e$ to GAN model to produce clean variant. 5) Filter `clean` OGM back into OGM format from $OGM_e$ and publish it as a map to the robot. 
    }
    \label{fig:network_diagram}
\end{figure*}

We extract generator model $G$ from a I2I GAN and place it into our SLAM process, this is is depicted as $G_{enc}/D_{enc}$ in Fig. \ref{fig:network_diagram}. We take the single input of incoming streams of 3D LIDAR data. The LiDAR is used to calculate the odometry as well as produce an estimated $OGM_e$. $OGM_e$ is preprocessed, and parsed to the generator model in which an output clean $OGM_c$ is predicted. This output is then processed back into the dimensions and occupancy values of $OGM_e$. The output is provided to the robot/operator. This entire process occurs in real-time.  

\subsection{Pose Estimation}
Our GAN model requires significant computational power, which can be limited within an embedded system. Therefore, we opt for an efficient and lightweight LiDAR-odometry estimation. Our GAN can also resolve the potential difference in error caused by a less accurate LiDAR-odometry calculation. We adapt and modify the LiDAR-odometry computation from the DLO algorithm \cite{b17} without IMU data to calculate 6-DOF motion between consecutive LiDAR frames. Their method is originally applied to 3D SLAM. We adapt it to work for 2D OGM through removing the mapping component and translating 3D Pointclouds to 2D LaserScan representations. 

We collect temporally adjacent scans of times $k$ and $k-1$ from a 360° 3D LiDAR scanner. We estimate pose through calculating the transformation between successive scans in a classical ICP approach. Each scan is sent through a process of filtering. We apply a box-filter and remove all voxels within $1\,\mathrm{m}^3$ from the origin of the pointcloud. This process attempts to eliminate data points that belong to the robot or scanning apparatus. We then down-sample the remaining data points to reduce computational complexity through applying a 3D Voxel Grid Filter with a resolution of $0.25\,\mathrm{m}$. Each remaining voxel corresponds to a cubic volume of $0.25\,\mathrm{m} \times 0.25\,\mathrm{m} \times 0.25\,\mathrm{m}$. The output of this is a filtered pointcloud of roughly $7{,}500$ voxels, for reference an input pointcloud is roughly $30{,}000$ voxels. This process reduces data complexity to improve computational efficiency and removes outliers which could be potentially damaging to the result.

With the processed pointcloud we perform scan to scan matching via Generalised-ICP to calculate a relative transform  $\hat{X}^{\mathcal{L}}_{k}$, in LiDAR coordinate system $\mathcal{L}$, from filtered source ($P^{s}_{k}$) and target ($P^{t}_{k}$) pointclouds. 

Target pointcloud $P^{t}_{k}$ is the $k-1$, from source pointcloud $P^{s}_{k}$, ($P^{t}_{k}$ = $P^{s}_{k-1})$. We calculate GICP residual error $\mathcal{E}$ from GICP:
\begin{align}
   \mathcal{E}\left(\mathbf{X}_k^{\mathcal{L}} \mathcal{P}_k^{\mathrm{s}}, \mathcal{P}_k^{\mathrm{t}}\right)=\sum_{i=1}^{N} d_i^{\top}\left(\mathcal{C}_{k, i}^{\mathrm{t}}+\mathbf{X}_k^{\mathcal{L}} \mathcal{C}_{k, i}^{\mathrm{s}} \mathbf{X}_k^{\mathcal{L}^{\top}}\right)^{-1} d_i
\label{eq:GICP}
\end{align}

To calculate relative transform:

\begin{align}
    \hat{\mathbf{X}}_k^{\mathcal{L}}=\underset{\mathbf{X}_k^{\mathcal{L}}}{\arg \min } \mathcal{E}\left(\mathbf{X}_k^{\mathcal{L}} \mathcal{P}_k^{\mathrm{s}}, \mathcal{P}_k^{\mathrm{t}}\right)
    \label{eq:rel_trans}
\end{align}
  
In which $d_i=p_i^{\mathrm{t}}-\mathbf{X}_k^{\mathcal{L}} p_i^{\mathrm{s}}, p_i^{\mathrm{s}} \in \mathcal{P}_k^{\mathrm{s}}, p_i^{\mathrm{t}} \in \mathcal{P}_k^{\mathrm{t}}, \forall i$ and $\mathcal{C}$ is an estimated covariance matrix. Initial guess $\tilde{\mathbf{X}}_k^{\mathcal{L}}$ is set to identity matrix $\mathbf{I}$.

A local submap $S_k$ is used to allow more globally consistent motion estimates. It is used to calculate an optimal transform to world coordinate system $\mathcal{W}$ with source point cloud $\mathcal{P}_k^{\mathrm{s}}$ 

\begin{align}
  \hat{\mathbf{X}}_k^{\mathcal{W}}=\underset{\mathbf{X}_k^{\mathcal{W}}}{\arg \min } \mathcal{E}\left(\mathbf{X}_k^{\mathcal{W}} \mathcal{P}_k^{\mathrm{s}}, \mathcal{S}_k\right) .  
  \label{eq:s2m}
\end{align}

The final output is the estimated pose within the world frame of the LiDAR scanner.
$\mathbf{X}_k^{\mathcal{W}}$

\subsection{GAN Model for OGM Cleaning}

A GAN model is built up of a minimum of two neural networks, a Generator $G$ and a discriminator $D$. GANs are successful in the task of I2I translation through the use of an adversarial loss:

\begin{equation}
\begin{aligned} 
\mathcal{L}_{GAN}(G, D, X, Y) = E_{y \sim Y} \log D(y) \\
+ E_{x \sim X} \log (1-D(G(x)))
\end{aligned} 
\end{equation}

This encourages the generator to produce samples that are indistinguishable from training data. 

We build our GAN model with one ResNet generator broken up into $G_{enc}$ and $G_{dec}$ and base its architecture off \cite{b22}. The summarised goal of $G$ is to take an erroneous OGM from domain $X$, and translate it to a pixel perfect variant, indistinguishable from data in domain $Y$, as judged by discriminator, $D$.

To learn this mapping function through training we utilise a self-supervised patchwise contrastive loss \cite{b23}. $G_{enc}$ compares patches between OGMs in domains $X$ and $Y$ to establish similar features. Patchwise contrastive loss can be defined as: 

\begin{equation}
\begin{aligned}
L_{\text{con}} = -\log \left[ \frac{\exp(q \cdot k^+/\tau)}{\exp(q \cdot k^+/\tau) + \sum_{i=1}^{N-1} \exp(q \cdot k^-/\tau)} \right]
\end{aligned}
\end{equation}

Anchor point $q$ is a randomly located feature in generated sample $G(i_x)$, $K^+$ mirrors this location in input image $i_x$. For each $K^+$ there are $N - 1$ negative samples ($K^-$). $\tau$ denotes temperature hyper parameter. The gradient of $L_{con}$ is applied to anchor point $q$ to learn a unilateral mapping function to translate regions of error within occupancy grid maps into pixel perfect variants. 

We decide to use a contrastive loss as opposed to a cycle-consistency loss, as our translation requires some geometric shifts for angular error correction. Many features of an erroneous OGM, such as the general location of occupied regions such as walls and obstacles, must also remain consistent through translation. 

The default usage for a patchwise contrastive loss is to randomly select features for $q$, $K^+$ and $K^-$. This is potentially inefficient as $L_{con}$ may impose inconsistent changes to generator $G$. Random selection of features takes no consideration into what reflects domain specific characteristics. To improve the feature selection process we incorporate a module outlined in \cite{b24}. This module selects features for cross domain comparison based on significance. 

The discriminator for our GAN model which provides the adversarial loss is a PatchGAN discriminator from \cite{b25}.

\begin{figure*}
    \centering
    \includegraphics[width=\linewidth]{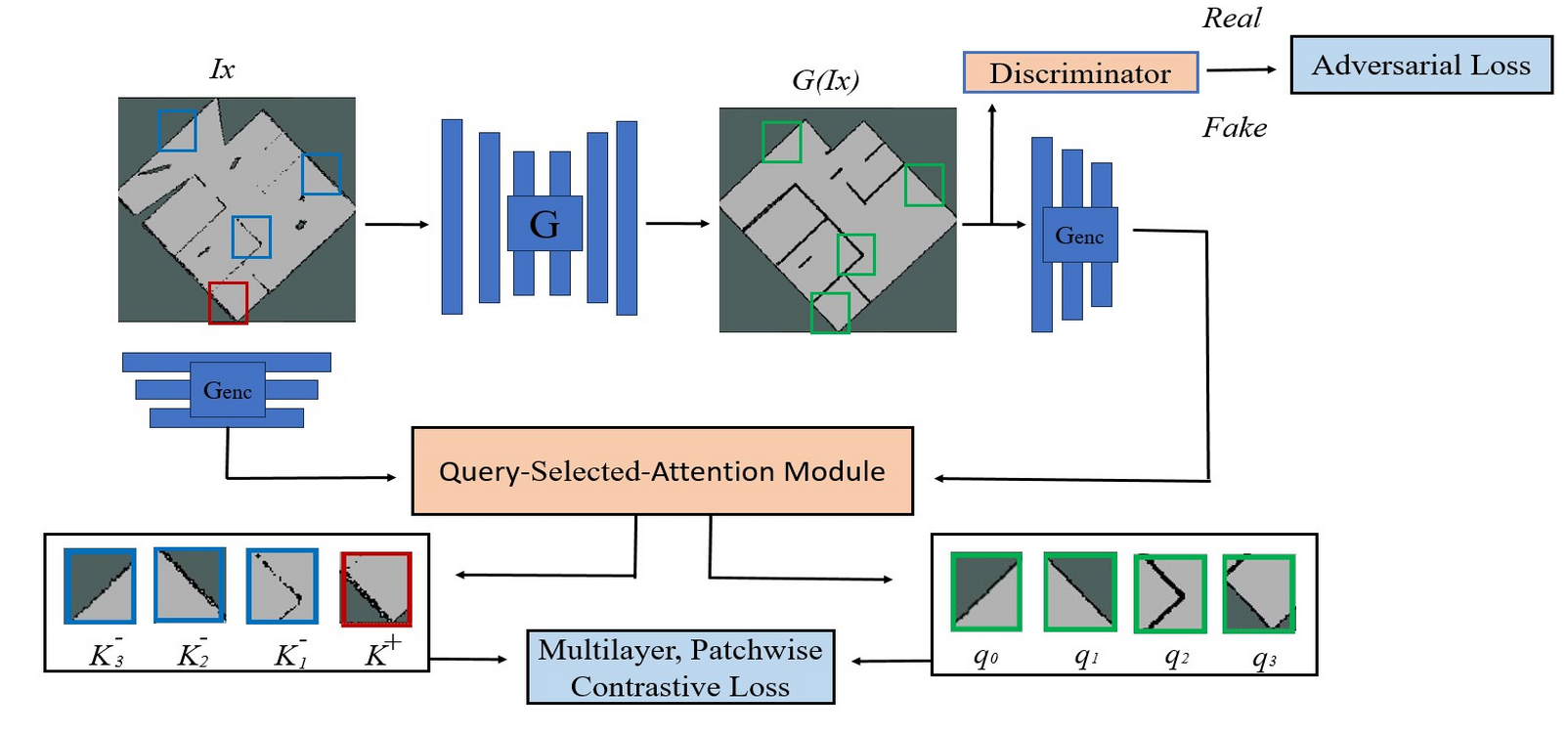}
    \caption{Training procedure for 2D occupancy grid cleaning through an I2I GAN, $I_x$ and $G(I_x)$ are generated by our modified PseudoSLAM simulator and are in image format, we use a Query-selected attention module \cite{b24} selection of $K^-$, $K^+$ and anchor points $q$ to compute a Multi-layer Patchwise Contrastive Loss from \cite{b23}.}
    \label{fig:model_diagram}
\end{figure*}

The full objective of our I2I GAN is:
\begin{equation}
\begin{aligned}
L_G = L_{\text{adv}} + L_{\text{con}}^X + L_{\text{con}}^Y
\end{aligned}
\end{equation}

where $L_\text{adv}$ is the adversarial loss from the PatchGAN discriminator. $L_{\text{con}}^X$ is the patchwise contrastive loss, and $L_{\text{con}}^Y$ is the identity loss. A diagram of our training process is available in Figure \ref{fig:model_diagram}

\subsubsection{Training Parameters}

The I2I GAN Model was trained for 100 epochs on a single Nvidia A100 GPU using PyTorch. Batch size was set to 1 for stability, learning rate was initialised as $2 \times 1^{-4}$ and linearly decays to 0. The Adam optimiser was used with $\beta =0.5$ and $\beta2 =0.999$. To track training progress, we use the Fréchet Inception Distance on 1000 samples kept out of training.
\subsection{Data}

To train a I2I GAN model a large amount of high quality data is required. For our desired mapping task we require several thousand image examples of erroneous occupancy grid maps and several thousand examples of pixel perfect grid maps. At the time of writing there exists no publicly available dataset of these domains. To build a dataset we take inspiration from the work presented in \cite{b26}. 

We use a deep reinforcement learning (DRL) agent trained for exploration to generate samples of erroneous maps. The agent computes a mimic of the Gmapping algorithm and incorporates errors into the output maps. We modified their baseline simulator into a data generation tool and run the agent in inference mode continuously to generate data. The agent explores environments of the HouseExpo2D dataset \cite{b26} and incorporates examples of common SLAM errors including: sensor noise, linear offsets, angular offsets, incomplete observations, and accidental observations. The agents is given a set amount of episodes in which it can move in the directions, forward/left/right. Reward is calculated based on total map explored in training. When the episode limit have been expended, the current explored environment is exported as an image and the agent is given a new map to explore. This process is split up between several agents at once. We created an erroneous copy of every sample within the HouseExpo 2D dataset (35,126 samples) through this process and use this as our domain $X$ for our I2I task. Domain $Y$ is the baseline HouseExpo 2D Dataset. 

\subsection{Simulator Setup}

\begin{figure}[h]
    \centering
    \includegraphics[width=0.8\columnwidth]{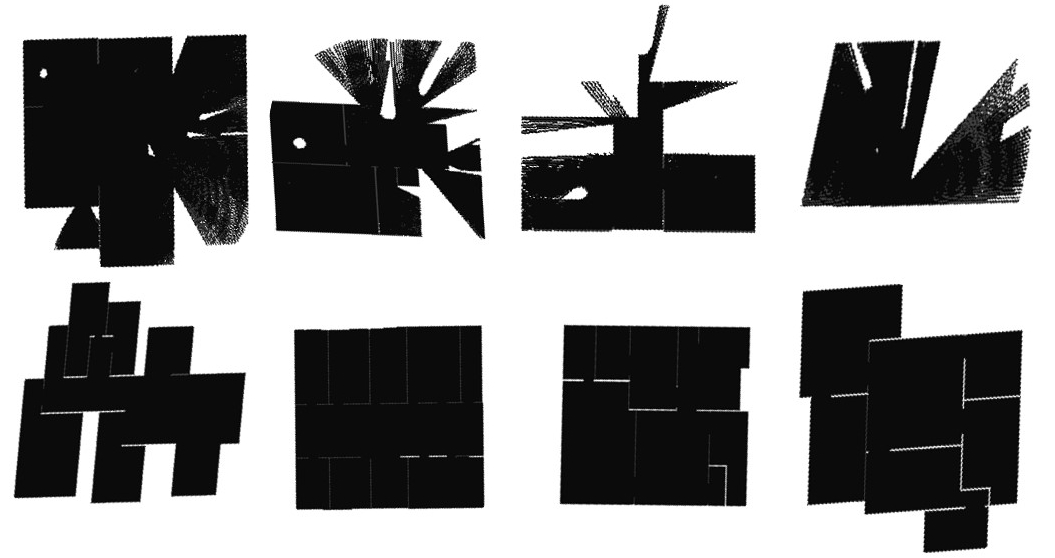}
    \caption{Samples from the dataset used to train the GAN model. Top row: samples of erroneous occupancy grid maps, bottom row: samples of `clean` occupancy grid maps.}
    \label{fig:drl_samples}
\end{figure}

The parameters for the final dataset for our model are available in Table \ref{tab:PseudoSLAM_parameters}. This dataset was built by 12 agents in total and once mapping was complete the data was augmented through random rotations and croppings increasing the final dataset amount to 89,344 samples. Rotating of samples is necessary for inference on real world data as occupancy grid maps are not cardinally align by default. To generate the entire dataset with these parameters split between 12 agents it took roughly 100 hours on a 10 core 2.90GHz CPU, the time usage is mostly consumed by the higher episode agents. The agents had increasing episode counts to generate samples with a variety of map completion, this ensures that models trained with the dataset are viable at every stage of map completion. The same applies to error severity, to ensure the model is appropriate for a variety of SLAM miscalculations and errors. Due to the nature of the maps being built through DRL, there is almost an unlimited amount of potential data that can be generated. The field of view (FOV) of the sensor used was set to 180\textdegree and the range set to a low value of 8 metres as it lead to increased map completion. This is due to the agent in the simulator observing gaps through open door ways which lowers the reward to explore further down these partially mapped pathways. With a higher range and FOV the agent would often become entrenched leading to lower map completion.

\begin{table}[h]
    \caption{PseudoSLAM parameters used in the modified simulator to produce data used to train our I2I GAN model.}
    \centering
    \includegraphics[width=0.8\columnwidth]{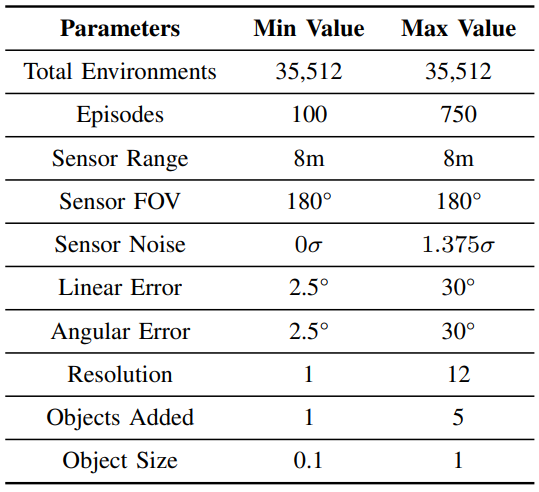}
    \label{tab:PseudoSLAM_parameters}
\end{table}

Samples of training data are available in Fig. \ref{fig:drl_samples}.

\subsection{GAN-SLAM}

Our algorithm takes input consecutive 3D LiDAR pointclouds $P(k, k+1, k..$) from a singular 360° LiDAR scanner, and outputs a high quality OGM $M$. To produce OGM $M$ we build a 2D data frame $LS_k$ of the 3D pointcloud through processing pointcloud data $P_k$. We extract Cartesian coordinates and intensity $x,y$ and $i$ from each point. We use $x, y$ to calculate Azimuth angle $\theta=atan2(y,x)$ and normalise output to be between $\pi $ and$ -\pi$. We calculate range of each point using the Euclidean distance representing distance from each point to the sensor. Each point is assigned to a specific bin within the 2D data scan based on its Azimuth angle. This new 2D representation mirrors the 3D pointcloud on which the 6 DOF egomotion of the scanner is calculated on. Therefore, aligning the odometry, calculated from the 3D Pointcloud, with the new 2D representation needed for OGM.  

We build estimated $OGM_e$ with the input of consecutive parallel 2D LiDAR $LS_k$ and pose estimate $\mathbf{X}_k^{\mathcal{W}}$. We incorporate the loop closure, map maintenance and other SLAM fundamentals from \cite{b21}. 
A Pose graph is constructed using odometry data plotting the trajectory over time, this data is also used for improving localisation accuracy within the OGM.

We preprocess input estimated $OGM_e$ through a simple process of filtering.
Grid cells are marked as outliers if they have a very low intensity and converted back to unexplored space. Cells with a very high intensity are rounded towards the closest value of occupation, $M_{ij} = 0$ or $M_{ij} = 1$. This process is done based on the idea of space confidence. Low intensity cells are interpreted as low confidence of occupation and offer little in value to the operator, these may be important for navigation tasks but are not useful for purely mapping. Whereas, cells of high intensity are important and are presented in the map as such. This process is only possible due to the implemented GAN model's ability to reconstruct the broken data. If applied to a traditional SLAM this process may cause issues. In this filtering many `good` pixels with low confidence are removed, such as some pixels within planar lines. This is not an issue for our SLAM as the GAN model is able to identify these, and effectively reconstruct the map based on the good data. 
This process of filtering is done at this stage rather than when converting the 3D $P_k$ to 2D $LS_k$ for efficiency. At this stage it does not have to be applied at the pointcloud level for every scan.

This processing should only be applied if the map is required for visual purposes. The values of grid cells hold significant importance in traditional evidence based OGM and can potentially cause damage to navigation tasks if altered in this manor. For this reason we advise our SLAM to be used for purely mapping based tasks. 

The output, filtered OGM is then taken through a process of floating point removal. Pixels that have two or less adjacent pixels of the same value are transformed back into unexplored cells. 

\begin{equation}
    Cell(x, y) = 
    \begin{cases}
        Cell(x, y) & \text{if } \sum_{i=1}^{4} \delta(p_i, C(x, y)) \geq 2 \\
        255 & \text{if }\sum_{i=1}^{4} \delta(p_i, C(x, y)) \leq 2 \\
    \end{cases}
\end{equation}

In which $p$ values represent neighboring cells and $\sum_{i=1}^{4} \delta(p_i, C(x, y))$ is a function that returns 1 if a chosen $p$ value is the same value as input $cell$, otherwise it returns 0. 255 is the value which represents unexplored cells.

The grid map is now represented with 3 distinct  values determining, occupied, unoccupied and unexplored cells (empty data). This is also done to improve the performance of the GAN model through matching the values within its training data. 

The exact dimensions ($H\times W$) of $OGM_e$ are saved, and it is then resized to 256x256, converted to a tensor and used as input for the generator $G$. The output clean $OGM_C$ is then mapped back to the original dimensions $H,W$ of the input $OGM_e$. $OGM_C$ is then processed through another filtering operation in which cells are rounded to their closest value of occupation. This is a necessary step when working between the data types of OGM and image. Slight changes in values within an image matrix can go un-noticed, however values of cells within an occupancy grid map carry much higher importance. 

The final output of this is OGM $M$. Trajectory, pose, orientation and other SLAM related data applies directly onto OGM $M$ from $OGM_e$. OGM $M$ can now be used directly for further downstream tasks.

\begin{figure}
   
        \includegraphics[width=\linewidth]{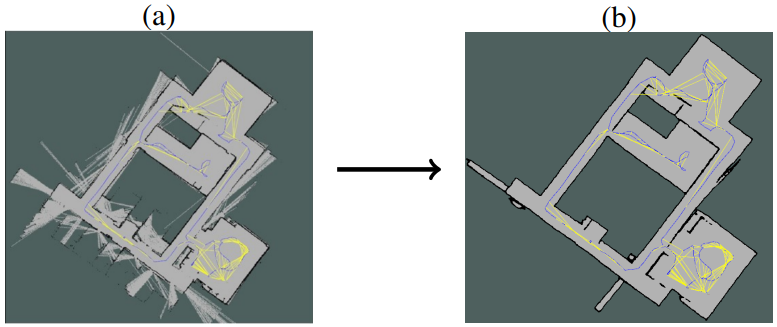}
   
    \label{fig:rtab_vs_gan-slam}
    \caption{A real-time prediction made by GAN-SLAM of the Haslegrave building at  
    REDACTED University. (a) Estimated OGM, used as input to GAN model, (b) Clean OGM variant prediction produced by GAN-SLAM.}
\end{figure}

\section{Results \& Discussion}

\subsection{Radish Dataset}

The Radish\cite{radish} dataset is a classic SLAM benchmark for 2D OGM. We use samples from the Radish dataset to demonstrate our models success at partial observation completion and error removal. We do not calculate odometry for pose estimation on these samples, but rather begin with an estimated OGM produced by Gmapping as provided by the dataset. We use this dataset to demonstrate the generalisability of our proposed method in a variety of large complex environments. Results are shown in Table \ref{tab:radish_new}.

\begin{table*}[ht]
    \centering
    \caption{Predictions made on samples from the Radish dataset (a) SDR Site B by Andrew Howards, (b) Fort AP Hill by Andrew Howards, (c) Logwood by Nick Roy (d) Intel Lab by Dieter Fox, (e) Albert-b Laser Vision (Freiburg)}
    \label{tab:radish_new}

        \includegraphics[width=2\columnwidth]{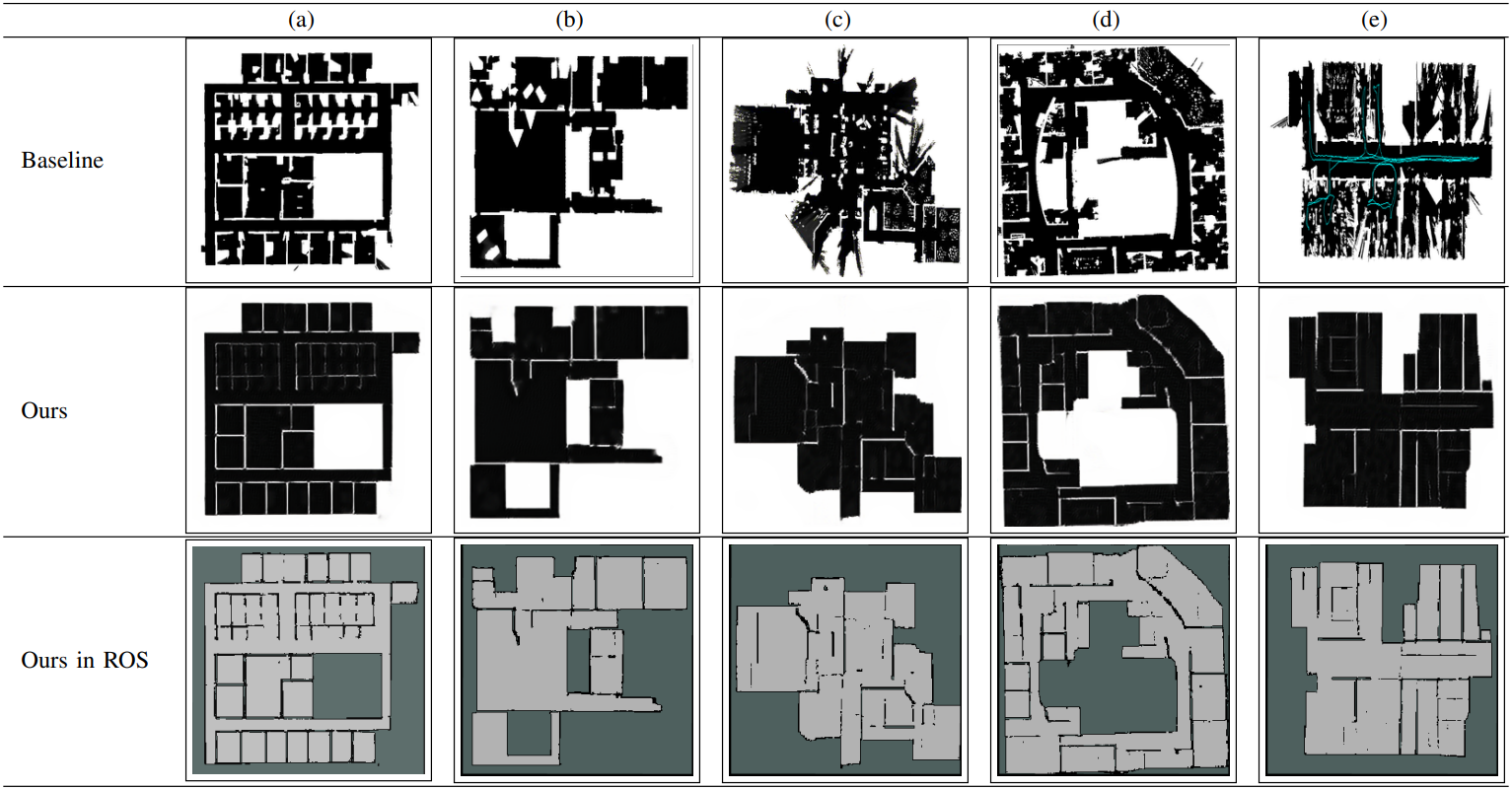}
        
        
\end{table*}

Our results for the Radish samples demonstrate clear success in partial observation completion and removal. This however was not perfect for every environment. In the Intel Lab SLAM map, our algorithm interpreted a corridor as an error, which could be problematic for path planning. We believe this occurred due to the resolution of the sample being compressed massively when it depicts quite a large environment. The corridor in question was reduced to only a couple of pixels in width which may have caused the GAN model to interpret it as an accidental observation. Metric evaluation for this criteria is a difficult task, as there exist no ground truth `completed` versions of samples in the Radish dataset. We believe our results demonstrate clear improvement over current baseline in terms of map completeness and invite readers to make their own conclusions.  

\subsection{Real-time SLAM}

To showcase the strength of our SLAM within the real world we record a complex mapping session of the Haslegrave building at REDACTED University with a Velodyne VLP-16 LiDAR scanner. We compare our results to other SLAM algorithms and evaluate them through IoU of occupied and unoccupied cell against a ground truth. We demonstrate all algorithms with and without our GAN model. To evaluate the performance of these SLAMs with our GAN, we use their methods to produce an OGM and use that as input to the GAN. Results are shown in Table.  \ref{tab:iou_comparison}. 

\begin{table}[H]
    \centering
    \caption{OGM Accuracy Comparison for Real-Time SLAM}
    \label{tab:iou_comparison}

        \includegraphics[width=1\columnwidth]{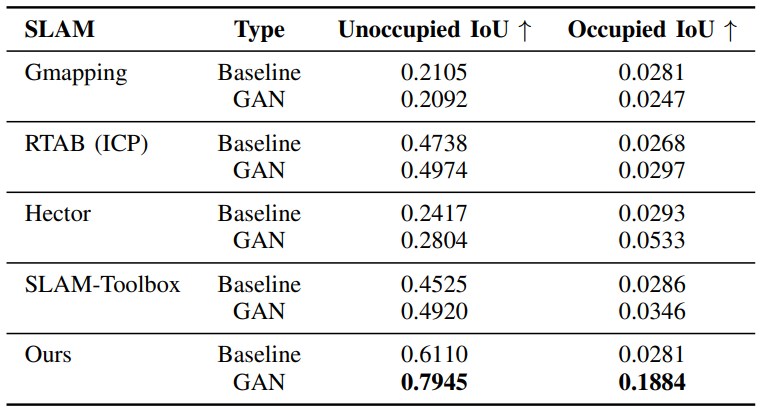}
        
\end{table}

The Haslegrave SLAM dataset was built to attempt to create several complex observations, such as instances of LiDAR going through glass and multiple sharp turns. We benchmark our algorithm against several well known 2D SLAM algorithms including: Gmapping \cite{b7}, RTAB-Map (with ICP Odometry) \cite{b21}, Hector SLAM \cite{b28}, and SLAM-Toolbox \cite{b29}. With these results we attempt to demonstrate that 2D LiDAR-odometry algorithms struggle to produce a viable map in a complex dynamic scene, whereas ours is able to. To produce these results we use the baseline implementation of each SLAM making no changes. The exact same dataset was used for all samples.

The ground truth sample used for this evaluation was manually created by an unaffiliated third party. We note that the occupied IoU is low for all SLAMs evaluated due to the discrepancy between ground truth and generated sample. In the ground truth only walls are marked as occupied space whereas in the SLAM every object is. We decide to show these results as despite this difference our GAN demonstrates significant improvement over baseline for all methods.

From our experimental testing we observe that all SLAM algorithms evaluated performed better with the usage of our GAN. We note that this accuracy is mainly appropriate and applicable for the context of purely mapping-based downstream tasks such as floor plan creation.

\begin{table}[h]
    \caption{Haslegrave SLAM Map Comparison}
    \centering

        \includegraphics[width=1\columnwidth]{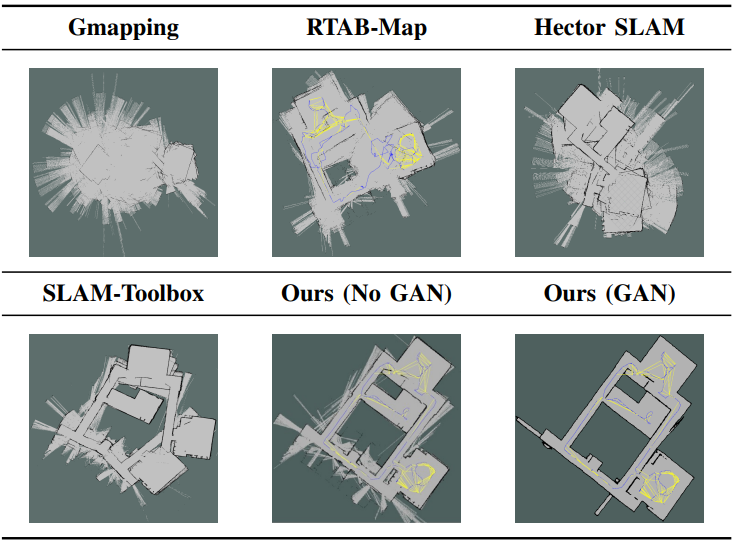}
    \label{tab:real-time-results}
\end{table}

\begin{figure}[h]
    \centering
    \caption{Ground Truth Map of the Haslegrave Area}
    \adjustbox{margin=3pt}{\includegraphics[width=0.25\textwidth]{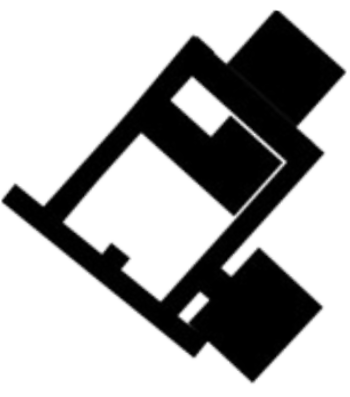}}
    \label{fig:ground_truth}
\end{figure}

\subsubsection{Computational Time Comparison}

Given that SLAM systems are often expected to run in real-time on robots with limited resources a potential criticism of our approach is the increase of computational time. To address this, we demonstrate computational time comparisons for all the aforementioned SLAM algorithms with and without our GAN model. Results are available in Table. \ref{tab:time_comparison}

\begin{table}[H]
    \centering
    \caption{Computational Time Comparison}
    \label{tab:time_comparison}

        \includegraphics[width=0.8\columnwidth]{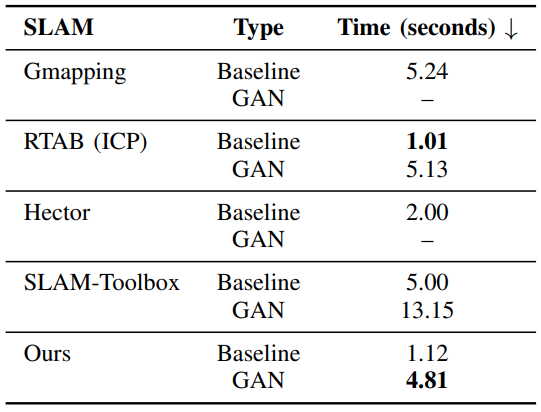}
\end{table}

From these results it can be observed that our method has a large increase in computational cost, this however is not too damaging and still is viable for real-time inference. Gmapping and Hector-SLAM were unable to operate in real-time with the GAN model, therefore results are not shown. 

For our experiments we use a 3D LiDAR as apposed to 2D to take advantage of the higher range. Our specific use-case is for floor plan creation within several thousand large factories across the UK in which the increase in range is necessary. We acknowledge that use of mature 2D Laser scan matching methods for 2D OGM may yield better results in smaller scenes, but at the cost of much lower range. This is unfeasible for large scale mapping tasks. 

\subsection{PseudoSLAM Simulated Data}

Due to the time taken to produce an evaluation of real world data considering that our use case focuses on large complex buildings. We also evaluate our SLAMs performance on simulated data. This is the evaluation data partition built by our DRL agent. To evaluate the performance of our SLAM, we compute the FID between predictions made by the trained generator on the 1000 samples of validation data, kept out of training. We compare these with a dataset of pixel perfect occupancy grid maps, also built by the simulator. We use FID as apposed to a more pixel comparison metric due to many of the samples being heavily incomplete with some rooms in the sample not being visited. Using pixel accuracy metrics would punish this heavily. We compare our GAN-SLAM with the baseline SLAM computed by the DRL agent, which attempts to mimic Gmapping. We note that we artificially introduce difficulties including sensor noise into the agent's baseline SLAM as the main purpose of this was to build samples of erroneous data. These results should not be interpreted as a evaluation or comparison of the simulators capabilities but rather a indication of how well our GAN-SLAM can improve these results. 

Samples and results are available for viewing in Figure \ref{tab:testing-images}. and Table \ref{tab:comparison_model_results}. Samples shown were chosen at random, the full 1000 predictions used for the FID calculation is available with our dataset.

\begin{table}[htbp]
    \centering
    \caption{Predictions made on samples of Testing Data (PseudoSLAM Simulator)}
    \label{tab:testing-images}

        \includegraphics[width=0.8\columnwidth]{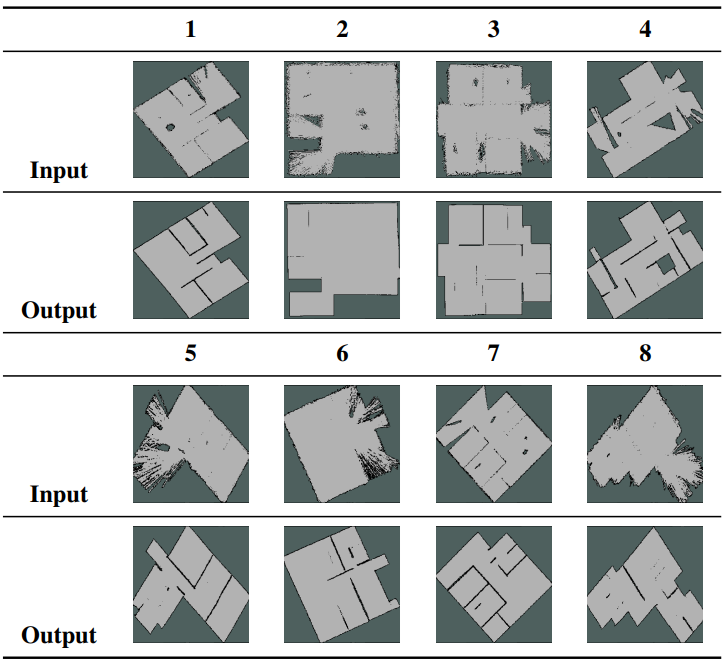}
\end{table}

\begin{table}[h!tb]
    \caption{Comparison of Model Results}
    \centering

        \includegraphics[width=0.4\columnwidth]{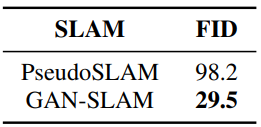}
    \label{tab:comparison_model_results}
\end{table}

\begin{figure*}[h]
   
        \includegraphics[width=\linewidth]{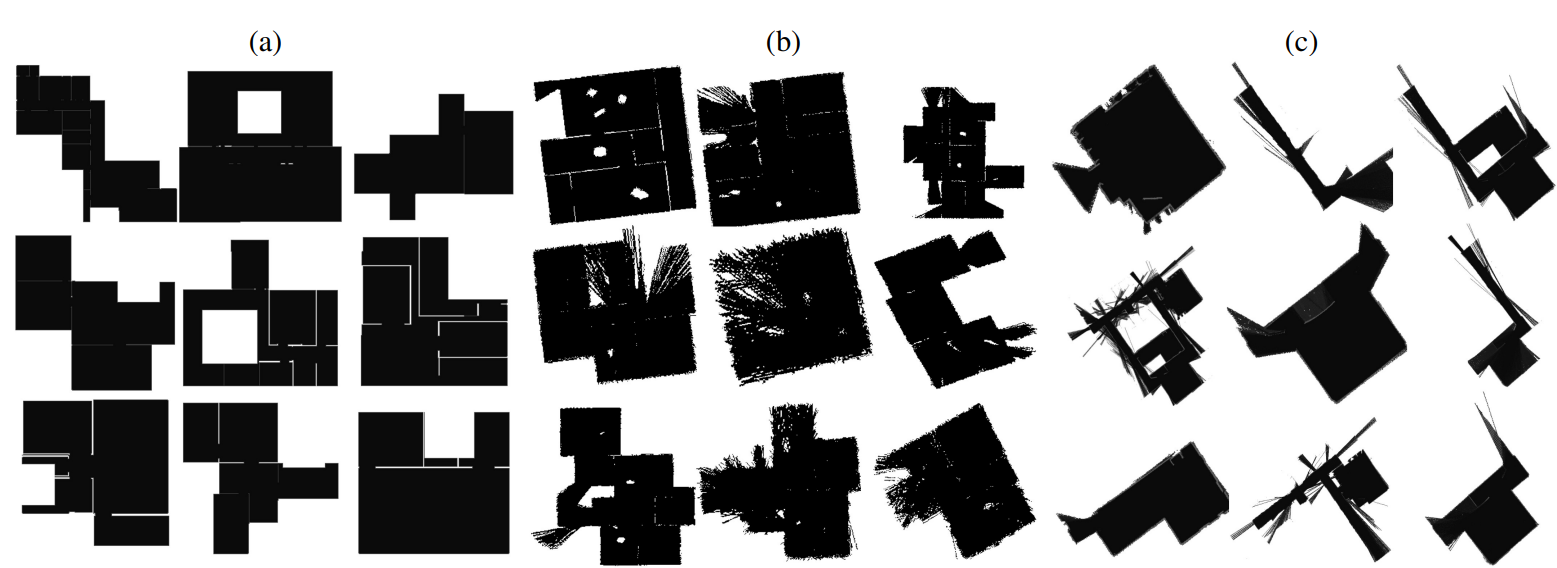}

        \caption{Comparison of data produced in the modified simulator and real world data. (a) Ground truth data generated within the modified simulator. (B) samples of erroneous occupancy grid maps generated in the modified simulator. (C) Samples of real data built by through occupancy grid mapping with a VLP-16 LiDAR Scanner in the Haslegrave building.}
    \label{fig:comparison}

\end{figure*}

\subsection{Modified Simulator Results}

 Our modified version of the PseudoSLAM simulator produced results with an average Structural Similarly Index Multiplier (SSIM) of 0.49 when compared to real data produced by RTAB-map in ROS2. The baseline simulator scores a SSIM result of 0.40.
 
 For these tests configuration of the exploration and mapping algorithm as well as the environments were kept the same. A comparison of samples produced by each simulator can be seen in Fig. \ref{fig:comparison}. Due to the nature of the maps being constructed by an DRL exploration algorithm, results of completion and error rate differed slightly between simulator results. To combat this mostly completed SLAM maps in the same environment were selected for comparison.

\subsection{Conclusions}

Our novel approach to GAN improved SLAM demonstrates robust and reliable mapping through a singular LiDAR scanner. Our SLAM integrates a new, previously unheard of, features such as room completion from partial observation and artefact removal in real-time. We demonstrate superior mapping quality as compared with other 2D OGM and 3D SLAM algorithms shown on real-world data. We believe our research demonstrates a potential new method of improving mapping quality in SLAM.

We acknowledge that using a generative model to modify data formats used for navigation tasks can be potential dangerous due to the chance of hallucinations. This is why we suggest only to use our method for purely mapping based tasked. 

Methods that we compare our SLAM to were chosen as they rely solely on LiDAR data to compute SLAM, which is how our algorithm operates. While there exist more accurate methodologies, these require usage of additional data such as IMU, or higher resolution LiDAR.

Our GAN-SLAM is ready to be used for large scale autonomous building surveying

\subsection{Limitations}

Due to our GAN model being trained on samples depicting mostly indoor buildings, the model generalises quite badly in outdoor scenes, often hallucinating incorrect layouts. Another area in which our model performs badly is within `rare` or obscure building designs, This is due to the majority of data depicting straight walls and `square` layouts, causing buildings with circular designs to be incorrectly hallucinated and mapped as square. A method to solve this limitation is to introduce a greater diversity of building designs within the training data.


\end{document}